# Video Text Localization using Wavelet and Shearlet Transforms

Purnendu Banerjee[a] and B. B. Chaudhuri[b*]

[a] Society for Natural Language Technology Research,
Module 130, SDF Building, Kolkata-700091, India
[b] Computer Vision and Pattern Recognition Unit,
Indian Statistical Institute, 203, B. T. Road, Kolkata-700108, India

## ABSTRACT

Text in video is useful and important in indexing and retrieving the video documents efficiently and accurately. In this paper, we present a new method of text detection using a combined dictionary consisting of wavelets and a recently introduced transform called shearlets. Wavelets provide optimally sparse expansion for point-like structures and shearlets provide optimally sparse expansions for curve-like structures. By combining these two features we have computed a high frequency sub-band to brighten the text part. Then K-means clustering is used for obtaining text pixels from the Standard Deviation (SD) of combined coefficient of wavelets and shearlets as well as the union of wavelets and shearlets features. Text parts are obtained by grouping neighboring regions based on geometric properties of the classified output frame of unsupervised K-means classification. The proposed method tested on a standard as well as newly collected database shows to be superior to some existing methods.

## 1. INTRODUCTION

With the growing popularity of digital video, the amount of available video data is increasing rapidly. Efficient navigation through such vast collections of digital video requires automatic indexing based on the content. For this task, algorithms are needed to automatically extract semantic information from the video.

One way of getting semantic information from the video is to employ the text contained in it. For example, captions in television newscasts summarize relevant names, locations and times pertaining to the news story. Movies and television programs include credits that indicate names of characters and actors. Extraction of text from video frames is more challenging than that from document images [1-2]. For example, Color video frames have much lower resolution than document images. Lossy video compression techniques (such as MPEG) cause color bleeding, blocking artefacts and loss of contrast. Also, Text stroke color is not known a priori, and the background is often complex and challenging [1-2] in the video.

The existing methods for text detection in camera based document images [6, 7] require high resolution and clear shape of the characters. Epshtein et al. [6] have proposed text detection based on stroke width transform. The stroke width transform works well if there is no disconnection in the character components. Pan et al. [7] have advocated a hybrid text detection approach based on Conditional Random Field (CRF), which involves connected component analysis to label the text. Another method of text extraction from camera images [8] works well if the text character shape is clear in the images. These constraints may be satisfied for high resolution scanned and camera images, but not necessarily for video based images, due to undesirable properties of video stated above. Thus, methods like [7, 8] may not be suitable for video frame images without suitable modifications.

At a high level, text in digital video can be grouped into two classes, namely scene text and graphics text. Scene text appears within the scene and is captured by the camera such as street signs, billboards, text on trucks, writing on shirts etc. Scene text is useful in many applications such as navigation, surveillance, video classification, or analysis of events [3-5]. Graphics text, on the other hand, is text that is mechanically added to video frames to supplement the visual and audio content. It is often more structured and closely related to the subject than scene text.

---
[*] Further author information:
Purnendu Banerjee: E-mail: purnendubannerjee@yahoo.com, Telephone: +91 9231427887
B. B. Chaudhuri.: E-mail: bbc@isical.ac.in, Telephone: (91) (33) 2575 2852

Text detection methods can be broadly classified into three categories, namely, connected component-based, edge and gradient based, and texture-based methods. The first category mainly uses geometric properties such as width and height of the connected components (CC) and intensity values. These methods are found to work well for horizontal graphics text [9-11]. But this approach does not work well for all video images because it assumes that text pixels in the same region have similar colors or grayscale intensities. The second category, based on edge and gradient methods, require text to have a reasonably high contrast with respect to the background. This category is not so good for low contrast frames in order to detect the edges. Also, the performance of such methods will deteriorate for too complex background in terms of false positives [12-17]. Another drawback of such techniques is the need for several threshold values at various stages.

The third category considers text as a special texture. These methods have used Fast Fourier Transform (FFT) [3], Discrete Cosine Transform, Wavelet Decomposition and Gabor filters for feature extraction. The texture based methods generally use classifiers such as Neural Network (NN), Support Vector Machine (SVM) etc. for text versus non-text classification [2, 18-19]. They have two types of drawback. First, NN and SVM require a large training set, sometimes in thousands of text/non-text samples. Also, it is hard to decide training set for non-text samples. Second, for large databases the texture-based methods are computationally expensive.

Among other methods, [20] has used entropy as feature and machine learning methods to identify the text. Such a method is good for high resolution image having background with nearly constant contrast, which may not be true for video text.

Identification of multi-oriented text has been partially addressed in [21, 22] where the algorithm is limited to location caption text at a few selected directions. Recently, Shivakumara et al. [23] have addressed this multi-oriented issue based on Laplacian and Skeletonization methods. That method works well for a video frame with different oriented text but not for a frame with non-linear curved text. A small study has been reported in [1] that extracts stylish text. This method again focuses on graphics and text region for limited directions. From these studies we can infer that more robust and effective method still needs to be developed in text detection and extraction.

In this paper, we propose a new text detection method executed by the following steps. (i) Compute Wavelet (Haar) and Shearlet transformation to enhance the text or high contrast information. (ii) Find Standard Deviation (SD) on combined value of two coefficients. The SD reflects the intensity variance and spatial gray value distribution of text and non-text [7, 14]. So, it is easier to classify text versus non-text pixels from the video frame. (iii) Employ K-means clustering to classify text pixels without prior training and supervision. (iv) Then the regions obtained by K-means clustering are merged by neighbourhood criteria and morphological operation to get the text zones.

## 2. BASICS OF SHEARLET

Wavelet is an advancement over frequency (Fourier) domain representation of signal where both time/space and frequency based representation are combined. It was initially proposed by Zweig in 1975 to study the reaction of the Cochlea of the ear to sound. To represent a signal, a set of basis function is employed in this formulation. One class of basis function is called Haar wavelet, but others also do exist.

Wavelet provides a kind of optimal approximation to one dimensional piecewise continuous function. However, they do not perform so optimally in 2D, especially in region of edges and other discontinuities. To handle this problem, new transforms like ridgelet, curvelet and contourlet, which need less number of basis functions to represent the discontinuity, have been proposed.

Another representation called Shearlet, obtained by translation, dilation and shearing transform on a finite state of generators that exhibit the desirable properties like multi-scale, localization, anisotropy and directionality, has been proposed more recently. This transform can be constructed using multi-resolution analysis and implemented efficiently by appropriate cascading algorithm. A comprehensive theory of completely supported shearlet is provided in [24, 25].

The problem of separating morphologically distinct features is underdetermined, since there is only one known data (the image) and two or more unknowns (features which are to be extracted from the known image). Experimental results using Morphological Component Analysis [38] suggest that such a separation task might be possible provided that we have prior information about the type of features to be extracted and the morphological difference between those is strong enough. Recently, it was theoretically proven that L1 minimization solves the separation of point and curve like features with arbitrarily high precision exploring a combined dictionary of wavelets and curvelets. Wavelets provide optimally sparse expansions for pointlike structures, and so does curvelets for curvelike structures. An associated algorithmic approach using wavelets and curvelets has been implemented in MCALab [28].

Shearlet systems are systems generated by one single generator with parabolic scaling, shearing, and translation operators applied to it, in the same way wavelet systems are dyadic scaling and translations of a single function, but including a directionality characteristic owing to the additional shearing operation (and the anisotropic scaling). Thus, a combined dictionary of wavelets and shearlets can be utilized for separating point-like and curve-like features. Moreover, numerical results give evidence to the superior behaviour of shearlet based decomposition algorithms when compared to curvelet based algorithms; see [29] for a comparison of ShearLab [30] with CurveLab [31].

The novel directional representation system of shearlets [32, 33] has emerged to provide efficient tools for analyzing the intrinsic geometrical features of a signal using anisotropic and directional window functions. In this approach, directionality is achieved by applying integer powers of a shear matrix, and those operations preserve the structure of the integer lattice which is crucial for digital implementations. This key idea leads to a unified treatment of the continuum as well as digital realm, while still providing optimally sparse approximations of anisotropic features. A formal introduction of shearlet systems in 2D is as follows

For $j \geq 0$ and $k \in \mathbb{Z}$, let

$$A_{2^j} = \begin{pmatrix} 2^j & 0 \\ 0 & 2^{\frac{j}{2}} \end{pmatrix}, \tilde{A}_{2^j} = \begin{pmatrix} 2^{\frac{j}{2}} & 0 \\ 0 & 2^j \end{pmatrix} \text{ and } S_k = \begin{pmatrix} 1 & k \\ 0 & 1 \end{pmatrix}$$

Next so-called cone-adapted discrete shearlet systems can be defined, where the term 'cone-adapted' is due to the fact that these systems tile the frequency domain in a cone-like fashion. Let c be a positive constant, which later controls the sampling density. For $\phi, \psi, \tilde{\psi} \in L^2(\mathbb{R}^2)$, the cone-adapted discrete shearlet system $SH(\phi, \psi, \tilde{\psi}; c)$ is defined by

$$SH(\varphi, \psi, \tilde{\psi}; c) = \Phi(\varphi; c) \cup \Psi(\psi; c) \cup \tilde{\Psi}(\tilde{\psi}; c),$$

Where

$$\Phi(\varphi; c) = \{\varphi(\cdot - cm) : m \in \mathbb{Z}^2\},$$

$$\Psi(\psi; c) = \{\psi_{j,k,m} = 2^{\frac{3j}{4}} \psi(S_k A_{2^j} \cdot - cm) : j \geq 0, |k| \leq \lceil 2^{j/2} \rceil, m \in \mathbb{Z}^2\},$$

$$\tilde{\Psi}(\tilde{\psi}; c) = \{\tilde{\psi}_{j,k,m} = 2^{\frac{3j}{4}} \tilde{\psi}(S_k^T \tilde{A}_{2^j} \cdot - cm) : j \geq 0, |k| \leq \lceil 2^{j/2} \rceil, m \in \mathbb{Z}^2\}.$$

In [34], a comprehensive theory of compactly supported shearlet frames having excellent spatial localization is given. Also, in [35] a large class of compactly supported shearlet frames were shown to provide optimally sparse approximations of images governed by curvilinear structures, in particular. These mechanisms are clearly explained in paper [26].

Here, to identify text in video images using wavelet and shearlet we follow the model of [26], which presumes that the image contains a collection of points and curve-like objects. In our case, we represent the input image as a one-dimensional (1-D) vector by simple reordering the two-dimensional (2-D) image to separate the text parts from the non-text parts. For such images $\mathcal{P}$ and $\mathcal{C}$ represent the point-like and curve-like components, respectively. Therefore, for any image $f$, we can write

$$f = \mathcal{P} + \mathcal{C} \tag{1}$$

Now, the Geometric Separation Problem consists of recovering $\mathcal{P}$ and $\mathcal{C}$ from the observed signal. It is known that the wavelets can provide optimally sparse approximation of signal which is smooth except at point discontinuities while shearlets provide sparse representation of curve-like structures. So, a composition of wavelet and band-limited shearlet can be used to expand the image function $f$.

Consider decomposition of $f$ into pieces corresponding to $j^{th}$ sub-band using filter $F_j$,

$$f_j = F_j * f \tag{2}$$

The Fourier transform $\hat{f}_j$ of the above function is supported on the scaling sub-band $j$ of the wavelet as well as the shearlet frame. Conversely, we can get,

$$f = \sum F_j * f_j, \quad f \in L^2(\mathbb{R}^2) \tag{3}$$

Let $\Phi_1$ and $\Phi_2$ be orthonormal bases of band-limited wavelets and a tight frame of band-limited shearlets respectively. Let, for a scale $j$ the image is separated into curve-like component $S_j$ and non-curve like component $W_j$. The best approximation to $S_j$ and $W_j$ can be obtained by solving the optimization expression,

$$(\widehat{W}_j, \hat{S}_j) = argmin_{W_j, S_j} \|\Phi_1^T W_j\|_1 + \|\Phi_2^T S_j\|_1 \quad \text{Where } f_j = W_j + S_j \tag{4}$$

It can be proved that,

$$\frac{\|\mathcal{P}_j - \widehat{W}_j\|_2 + \|\mathcal{C}_j - \hat{S}_j\|_2}{\|\mathcal{P}_j\|_2 + \|\mathcal{C}_j\|_2} \to 0, \quad j \to \infty \text{ Where } \|\hat{x}\|_i \text{ denotes the } L_i \text{ norm of the vector } \hat{x}.$$

In other words, $\mathcal{P}_j$ and $\mathcal{C}_j$ can be recovered with arbitrary high precision at very fine scale. Here, the image is decomposed clearly into point-like and curve-like parts. But for an arbitrary image, this assumption is not true, as it may also contain noise that is not represented well both by $W_j$ and $S_j$. It is very hard to propose a dictionary that leads to sparse representation for a wide family of signal. Therefore, the solution for this problem could be obtained by relaxing the constraint in equation (4) as follows,

$$(\widehat{W}_j, \hat{S}_j) = argmin_{W_j, S_j} \|\Phi_1^T W_j\|_1 + \|\Phi_2^T S_j\|_1 + \lambda \|f_j - W_j - S_j\|_2^2 \tag{5}$$

In this new form, the additional content in the image – the noise – characterized by the property that it cannot be represented sparsely by either one of the two representation systems, is allocated to the residual $f_j - W_j - S_j$. In this we not only mange to separate both point and curve-like components, but also succeed in removing an additive noise as a by-product. Now, by performing this minimization, we can separate point and curve-like objects, which were modelled by $\mathcal{P}_j$ and $\mathcal{C}_j$, and remove an additive noise component. However, solving equation (5) for all relevant scales $j$ is computationally expensive.

Instead of reconstructing $f$ using equation (3), an approximate reconstruction $\hat{f}_j$ is obtained by using weights $W_j$ at each sub-band. The weight is very small at zero or low sub-band and higher at higher sub-band. Following Lim et al. [26], we have considered four sub-bands in our problem. Here the weights given in section 4 are chosen by testing on a large number of images.

## 3. TEXT LOCALIZATION APPROACH

Our proposed text segmentation method contains two subsections: Shearlet-Wavelet features computations and then text area boundary refinement. The application of Shearlet-Wavelet on the input image to extract the features is described below.

### 3.1 Shearlet- Wavelet Feature Computations

The image can be decomposed into one sub-bands representing low frequency components (coarse scale), several sub-bands representing high frequency components (detailed scales), and one sub-band representing the very high frequency components (fine scale). Text in images is mostly represented by the high frequency components. Hence the low frequency sub-band representing non-text region is ignored in our experiment. The sub-band of very high frequency components that mostly represent noise is also ignored.

For the Shearlet transform, we make 2D convolution of the image with discretized band-limited shearlets, first introduced in [27]. In this implementation the directional selectivity of shearlets can be faithfully adapted to the digital setting. This allows precise extraction of directional features, e.g., edges, which is crucial for extracting curve-like components. It has been shown that for image separation, band-limited shearlets are superior to compactly supported shearlets, since they provide a tight frame as well as excellent directional selectivity due to high localization in the frequency domain.

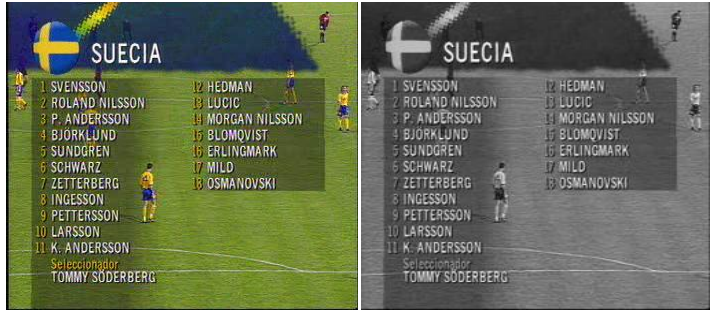

Figure 1: Original Color Image (Left) Gray Image (Right).

In our approach, the images are converted from color to gray (See Figure 1) and resized into 256 x 256 pixels [10]. Both shearlet and wavelet decompositions are done on this gray tone image. The high frequency sub-bands provide high shearlet coefficients for the text edge pixels. Similarly, wavelets provide high coefficient values for the pixels inside the body of the text.

Our algorithm progresses in an iterative manner. At first, we calculate the shearlet coefficient ($\widehat{S_k}$) and wavelet coefficient ($\widehat{W_k}$) for the k$^{th}$ iteration. After the first iteration, we take summation of both $\widehat{S_k}$ and $\widehat{W_k}$ coefficient values in an image $I_k$ and subtract from the original image, while residual valued image is sent to the next iteration for further processing. In this way, we have continued five iterations and the final results are shown in pseudo color in Figure 2. The pseudo color is generated by quantizing the co-efficient value in the 256 discrete ranges between blue to red, where the blue color is mapped to the lowest value 0 and 255 is red. Here, in Figure 2(a) the border of text is prominent (as the lines in bottom of the image are visible), while in Figure 2(b) the inside portion of the text is prominent (hence the bottom lines are absent). The combined image contains more complete description of the text but also the lines, which are removed by subsequent clustering.

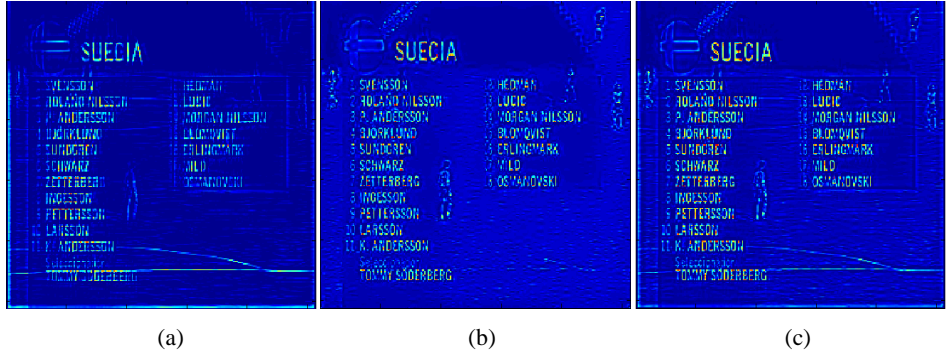

(a)          (b)          (c)

Figure 2: The Final Coefficient Images where (a) Final Shearlet Image, (b) Final Wavelet Image and (c) Final Combined Image.

In our approach, we compute Standard Deviation (SD) of the result of last iteration ($I$). The standard deviation of the combined valued representation $I$ in a sliding $3 \times 3$ window at position $(x, y)$ is defined as:

$$\sigma_{win}(x, y) = \sqrt{\frac{1}{N^2} \sum_{i=-1}^{+1} \sum_{j=-1}^{+1} (I(x+i, y+j) - \bar{I}(x,y))^2} \tag{6}$$

where $I(x+i, y+j)$ is the combined value of shearlet and wavelet coefficients at pixel position $(x+i, y+j)$, and the average $\bar{I}(x,y)$ is defined as:

$$\bar{I}(x,y) = \frac{1}{N^2} \sum_{i=-1}^{+1} \sum_{j=-1}^{+1} I(x+i, y+j) \tag{7}$$

## 3.2 Boundary Refinement

It is difficult to determine the boundary of text blocks directly from the text cluster because of false positives and disconnected text (Figure 3(a)). The regions obtained from K-means segmentation are in small piece and contain both text and non-text regions. After segmentation of text and non-text regions, we group the small regions together based on the abundance of text pixels to get bigger connected components as follows.

We consider a N×N window where N = 7 for every pixel and for each window position we compute the Text Presence Ratio ($TPR$). Here we decide the window containing the text components or not by relative abundance on K-means clustered image. We define,

$$TPR = \frac{No.\,of\,text\,pixels\,in\,the\,window}{Total\,no.\,of\,pixels\,in\,the\,window} \quad (8)$$

Given a threshold $T$, if $TPR < T$, the candidate window is considered as a false positive (shown white); otherwise, it is accepted as true text block (shown black). The resulting text regions are further smoothed by using a $3 \times 3$ morphological Closing operation. The result obtained after this operation on Figure 3(a) is shown in Figure 3(b).

The result is fed to K-means based clustering to classify text and non-text pixels, using on the features $\sigma_{win}(x,y)$ and corresponding I(x,y) values. The K-means clustering is unsupervised and we use the higher value of combined coefficients as the criteria to decide for text cluster and the lower values for the non-text cluster. A typical result of K-means clustering on Figure 2(c) is shown in Figure 3(a).

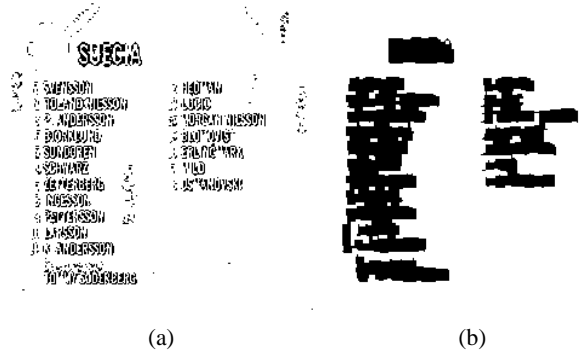

(a)          (b)

Figure 3: K-means clustered Image (a) and Boundary Redefined Image (b).

## 4. EXPERIMENTAL RESULTS

As there is no standard benchmarking dataset available, we have selected 140 video frames, extracted from TV news programmes, sports videos and movie clips. In addition to our own dataset, we consider another publicly available datasets of 45 video frames from National Laboratory on Machine Perception Peking University and Microsoft Research China (http://www.cs.cityu.edu.hk/~liuwy/PE_VTDetect), in which 36 frames are from Spanish TV RTVE, and 9 are from Ministry of Education, Singapore. The image sizes range from 320× 240 to 640 × 360 pixels. The parameter values are experimentally determined: N = 7, T = 0.5 and weights are 0.1, 0.1, 1.5 and 1.5 for low frequency to high frequency, respectively.

For comparison with our approach, we have implemented three existing methods described in [11, 5 and 16]. The method in [11], denoted as edge-based method, extracts edge features by using the Sobel operator. The method of [5], denoted as uniform-colored method, performs clustering in the L*a*b color space to locate the text lines. Finally, the method [16], denoted as Laplacian-based method, has clustered the image by K-means based on the gradient difference value for each pixel in the Laplacian-filtered image.

Figure 4 shows two sample results by these three methods and our proposed method. Image 4(a) has two low contrast text blocks. The edge-based method fails to detect any text block because of the problem of fixing threshold values for edge detection. The uniform-colored method detects the text blocks with missing characters and produces many false positives due to the problem of color bleeding. The method based on Laplacian and Skeletonization has detected the text blocks with excess area. In comparison, the proposed method detects the blocks more accurately.

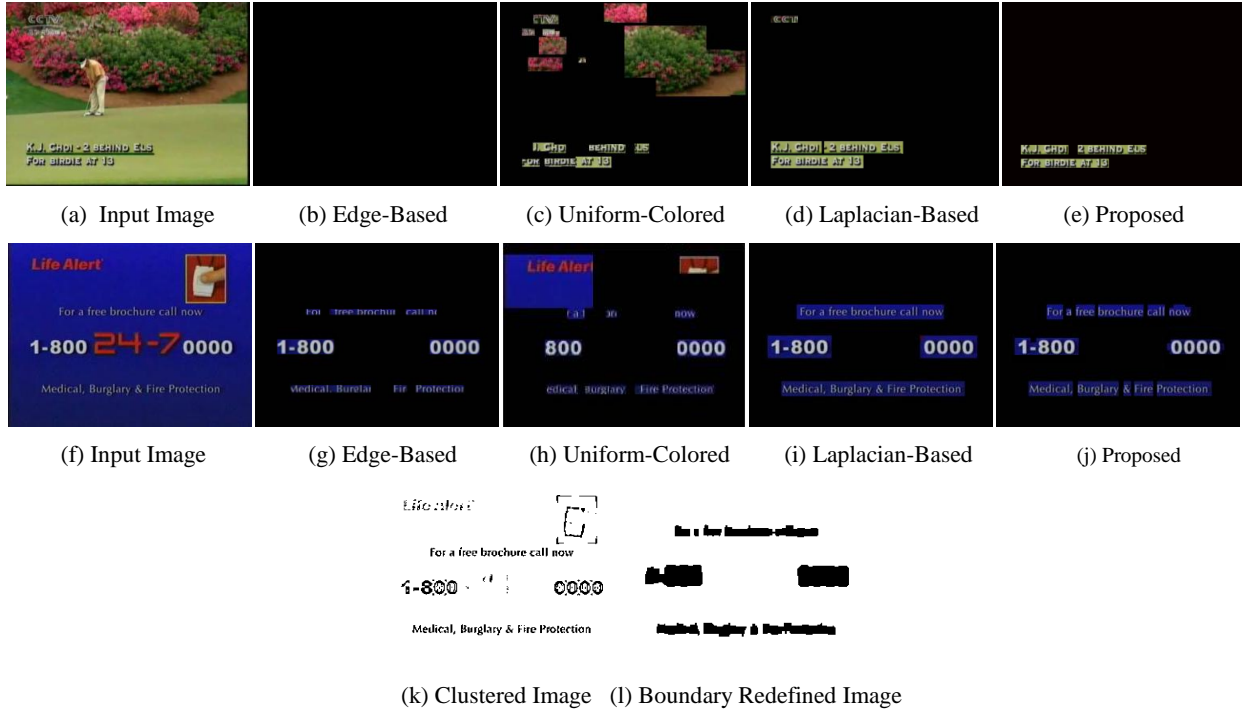

(k) Clustered Image   (l) Boundary Redefined Image

Figure 4: The detected text blocks by three existing methods and the proposed method for input images (a) and (f).

In Figure 4(f) an image is shown where the proposed method fails to detect some text blocks like "Life Alert" and "24-7" in red color. The "Life Alert" is partly there in our clustered image, but the another text part has vanished (see Fig. 4(k)). In the boundary refinement process the partly visible signature is completely wiped out (see Fig. 4(l)). The Laplacian-based and edge-based methods also could not detect the same text blocks, as shown in Fig. 4(i) and 4(g). On the other hand, uniform-colored method is able to detect one of the red "Life Alert" text blocks, but also detects some non-text block as text block on the right top corner of Fig. 4(h).

### 4.1 Evaluation on the dataset

There are several way to represent the evaluation result like as Shivakumara et al. [16], define the categories for each detected block by a text detection method given by Detection Rate (DR) = TDB / ATB, False Positive Rate (FPR) = FDB/(TDB + FDB), Misdetection Rate (MDR) = MDB / TDB where, Actual Text Blocks (ATB) is counted manually and other acronyms are as follows.

>True detected Block (TDB): A detected block that contains a text, partially or fully.

>False Detected Block (FDB): A detected block that does not contain text.

>Text Block with Missing Data (MDB): A detected block that misses some characters of a text block (MDB is a subset of TDB).

In this paper, to compare our method with other published algorithm, measure of [10] is used. The output of each algorithm is a set of rectangles designating bounding boxes for detected text region. This set is called the estimate. A set of ground truth boxes, called the targets is provided in the dataset. The match $m$ between two rectangles is defined as the area of intersection divided by the area of the minimum bounding box containing both rectangles. This number has the value 1 for identical rectangles and 0 for rectangles that have no intersection. For each estimated rectangle, the closest match was found in the set of targets, and vice versa. Hence, the best match $m(r, R)$ for a rectangle $r$ in a set of rectangle $R$ is defined by

$$m(r, R) = max\{m(r, r')|r' \in R\} \qquad (9)$$

Then, the definitions for recall and precision is

$$Recall(R) = \frac{\sum_{r_t \in T} m(r_t, E)}{|T|} \quad (10)$$

$$Precision(P) = \frac{\sum_{r_e \in E} m(r_e, T)}{|E|} \quad (11)$$

where $E$ and $T$ are the estimated and ground truth rectangles respectively.

The standard f-measure (f) was used to combine the precision and recall figures into a single measure of quality. The relative weights of these are controlled by a parameter $\alpha$, which we set to 0.5 to give equal weight to precision and recall:

$$f = \frac{1}{\frac{\alpha}{P} + \frac{(1-\alpha)}{R}} \quad (12)$$

Tables 1 show the performance of the three existing methods and the proposed method on both kind of dataset. The proposed method has the highest recall and f-measure value on video database. It outperforms the edge-based method and the uniform-colored method in all the performance measures. Compared to the Laplacian-based method, the proposed method has better recall and f-measure but worse precision. However, slightly higher precision might be compensated by the significant difference in recall between the two methods. On the whole, the proposed method has achieved better detection results on the database.

Table 1: Results on the video database (in %).

| Method | Recall($R$) | Precision($P$) | f-measure($f$) |
|---|---|---|---|
| Edge-Based [11] | 76.94 | 80.35 | 78.61 |
| Uniform-colored [5] | 52.58 | 73.26 | 61.22 |
| Laplacian-Based [16] | 84.38 | **86.95** | 85.66 |
| Proposed | **87.23** | 85.64 | **86.43** |

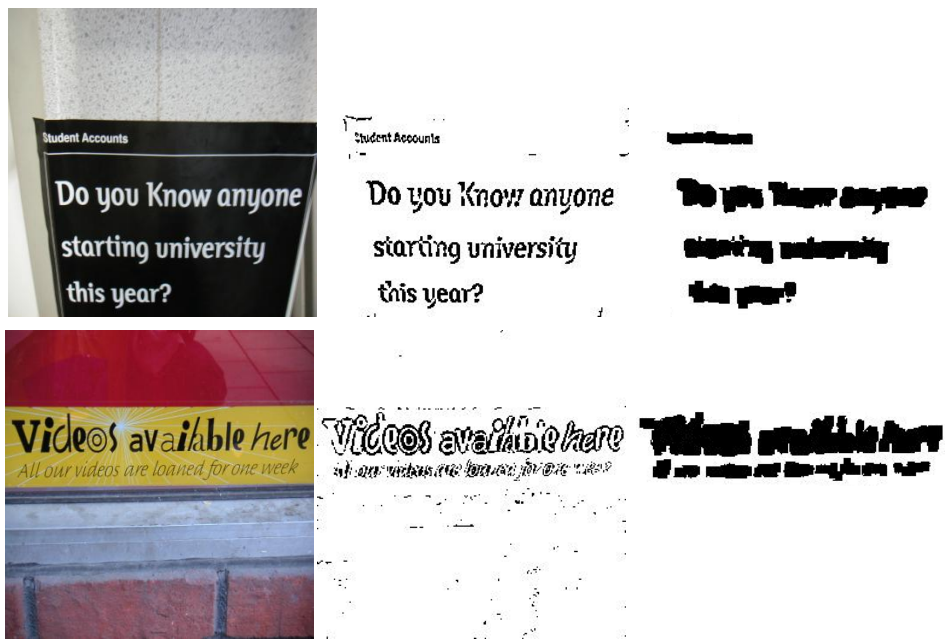

Figure 5: Result of Clustered and Boundary Redefining stage on test images from ICDAR 2011 competition dataset

On the other hand, to examine the robustness of our algorithm we employ our method on scene images the dataset that is provided from the ICDAR 2011 Robust Reading Competition [37]. Note that the approach works fairly well on scene image as well. Scene text detection result of two test images from ICDAR 2011 competition dataset are given in Figure 5.

## 5. CONCLUSION

We have proposed an efficient method for text detection based on the combination of Shearlet and Wavelet coefficients. An application of the text detection approach on Bangla and Devanagari Video Text was explained in [39]. Here, the gradient information helps to identify the candidate text regions and the edge information serves to determine the accurate boundary of each text block. Experimental results show that the proposed method outperforms the three existing methods in terms of detection and false positive rates. In the future, we will incorporate context information into the system, especially with machine learning method. Besides, the full system is implemented under Matlab 2010a and usually takes several seconds to process a normal size image on a PC with Intel(R) Core (TM) 2 Duo CPU 3.00 GHz.

## ACKNOWLEDGMENTS

This research is partly supported by Indian Statistical Institute and Society for Natural Language Technology Research, Kolkata.

## REFERENCES

1. D. Crandall and R. Kasturi, "Robust Detection of Stylized Text Events in Digital Video", in Proc. ICDAR, pp. 865-869, 2001.
2. K. Jung , "Neural network-based text location in color images", Pattern Recognition Letters, vol. 22, pp. 1503-1515, 2001.
3. J. Zang and R. Kasturi, "Extraction of Text Objects in Video Documents: Recent Progress", in Proc. DAS, pp. 5-17, 2008.
4. K. Jung, K.I. Kim and A.K. Jain, "Text Information Extraction in Images and Video: a Survey", Pattern Recognition, vol. 37, pp. 977-997, 2004.
5. V. Y. Mariano and R. Kasturi, "Locating Uniform-Colored Text in Video Frames", in Proc. ICPR, pp. 539-542, 2000.
6. B. Epshtein, E. Ofek and Y. Wexler, "Detecting Text in Natural Scenes with Stroke Width Transform", in Proc. CVPR, pp. 2963-2970, 2010.
7. Y. F. Pan, X. Hou and C. L. Liu, "A Hybrid Approach to Detect and Localize Texts in Natural Scene Images", IEEE Trans. on Image Processing, vol. 20, pp. 800-813, 2011.
8. P. P .Roy, U. Pal, J. Llados and F. Kimura, "Multi-Oriented English Text Line Extraction using Background and Foreground Information", in Proc. DAS, pp. 315-322, 2008.
9. A.K. Jain and B. Yu., "Automatic Text Location in Images and Video Frames", Pattern Recognition, vol. 31, pp. 2055-2076, 1998.
10. Q. Ye., Q. Huang, W. Gao and D. Zhao., "Fast and robust text detection in images and video frames", Image and Vision Computing, vol. 23, pp. 565-576, 2005.
11. K. Jung and J. H. Han, "Hybrid Approach to Efficient Text Extraction in Complex Color Images", Pattern Recognition Letters, vol. 25, pp. 679–699, 2004.
12. C. Liu, C. Wang and R. Dai., "Text Detection in Images Based on Unsupervised Classification of Edge-based Features", in Proc. ICDAR, pp. 610-614, 2005.
13. P. Shivakumara, W. Huang and C.L. Tan., "An Efficient Edge based Technique for Text Detection in Video Frames", in Proc. DAS, pp. 307-314, 2008.
14. M. Cai, J. Song and M. R. Lyu, "A New Approach for Video Text Detection", in Proc. ICIP, pp. 117-120, 2002.
15. E. K. Wong and M. Chen., "A new robust algorithm for video text extraction", Pattern Recognition, vol. 36, pp. 1397-1406, 2003.
16. T. Q. Phan, P. Shivakumara and C. L Tan, "A Laplacian Method for Video Text Detection", in Proc. ICDAR, pp. 66-70, 2009.
17. N. Sharma, P. Shivakumara, U. Pal, M. Blumenstein and C. L. Tan, "A New Method for Arbitrarily-Oriented Text Detection in Video", in Proc. DAS, pp. 74-78, 2012.

18. H. Li, D. Doermann and O. Kia, "Automatic Text Detection and Tracking in Digital Video", IEEE Transactions on Image Processing, vol. 9, pp. 147-156, 2000.
19. P. Shivakumara, T. Q. Phanand C. L Tan, "A Robust Wavelet Transform Based Technique for Video Text Detection", in Proc. ICDAR, pp. 1285-1289, 2009.
20. X. Chen and A. L . Yuille, "Detecting and Reading Text in Natural Scenes", in Proc. CVPR, pp. 366-373, 2004.
21. H. Tran, A. Lux, H. L. T. Nguyen and A. Boucher, "A Novel Approach for Text Detection in Images using Structural Features", in Proc. ICAPR, pp. 627-635, 2005.
22. J. Zhou, L. Xu, B. Xiao and R. Dai, "A Robust System for Text Extraction in Video", in Proc. ICMV, pp. 119-124, 2007.
23. P. Shivakumara, T, Q. Phan and C. L. Tan, "A Laplacian Approach to Multi-Oriented Text Detection in Video", IEEE Trans. on PAMI, vol. 33, pp. 412-419, 2011.
24. P. Kittipoom, G. Kutyniok, and Wang-Q Lim, "Construction of compactly supported shearlet frames", Constructive Approximation, Springer, vol. 35, pp. 21-72, 2012.
25. G. Kutyniok J. Lemvig and Wang-Q Lim, "Shearlets are Optimally Sparse Approximations", Applied and Numerical Harmonic Analysis, pp. 145-197, 2012.
26. G. Kutyniok and Wang-Q Lim, "Image Separation using Wavelets and Shearlets", in Curve and Surfaces, Springer LNCS, vol. 6920, pp. 416-430, 2012.
27. G. Easley, D. Labate, and Wang-Q Lim, "Sparse directional image representations using the discrete shearlet transform", Applied and Computational Harmonic Analysis, vol. 25, pp. 25–46, 2008.
28. MCALab (Version 120) is available from http://jstarck.free.fr/jstarck/Home.html
29. G. Kutyniok, J. Lemvig, and W.-Q Lim, "Compactly Supported Shearlets", Approximation Theory XIII: San Antonio 2010, Springer Proc. In Mathematics, vol. 13, pp. 163-186, 2012.
30. ShearLab (Version 1.0) is available from http://www.shearlab.org.
31. CurveLab (Version 2.1.2) is available from http://www.curvelet.org.
32. D. Labate, W.-Q Lim, G. Kutyniok, and G. Weiss, "Sparse multidimensional representation using shearlets", in Wavelets XI, SPIE Proc. Bellingham, WA, vol. 5914, pp. 254–262, 2005.
33. K. Guo, G. Kutyniok, and D. Labate, "Sparse multidimensional representations using anisotropic dilation and shear operators", Wavelets and Splines, Nashboro Press, Nashville, TN, pp. 189–201, 2006.
34. P. Kittipoom, G. Kutyniok, and W.-Q Lim, "Construction of compactly supported shearlet frames", Construction Approximation, vol. 35, pp. 21-72, 2012.
35. G. Kutyniok and W.-Q Lim, "Compactly supported shearlets are optimally sparse", Journal of Approximation Theory, vol. 163, pp. 1564-1589, 2011.
36. E. J. Candes and D. L. Donoho, "New tight frames of curvelets and optimal representations of objects with piecewise C2 singularities", Communication on Pure and Applied Mathematics, vol. 57, pp. 219–266, 2004.
37. A. Shahab, F. Shafait, and A. Dengel, "ICDAR 2011 robust reading competition challenge 2: Reading text in scene images", in Proc. ICDAR, pp. 1491–1496, 2011.
38. J. L. Starck, M. Elad, and D. Donoho, "Image decomposition via the combination of sparse representation and a variational approach", in Proc. IEEE Trans. Image Processing, vol. 14, pp. 1570–1582, 2005.
39. P. Banerjee and B. B. Chaudhuri, "An Approach for Bangla and Devanagari Video Text Recognition", In International Workshop on Multilingual OCR (MOCR), Washington DC, USA, 2013.